\documentclass{llncs}
\usepackage{wrapfig}
\usepackage{amsmath, tikz, url}
\usetikzlibrary{fit,positioning}
\usepackage{booktabs,tabu}
\usepackage{bm}
\usepackage{amssymb}
\usepackage{array}
\usepackage{multirow}
\usepackage{subfig}
\usepackage[]{algorithm2e}
\usepackage{wrapfig}

\allowdisplaybreaks
\newcolumntype{C}{@{\extracolsep{.5cm}}c@{\extracolsep{7pt}}}%
\usepackage{subfig}

\newcommand{\ie}{i.e.}
\DeclareMathOperator{\tr}{tr}
\begin{document}
\pagestyle{headings}  

\title{Cross-lingual Document Retrieval using Regularized Wasserstein Distance}

\author{Georgios Balikas\inst{1} \and
Charlotte Laclau\inst{1}\and
Ievgen Redko\inst{2}\and
Massih-Reza Amini\inst{1} 
}
\authorrunning{G. Balikas et al.} 

\institute{Univ. Grenoble Alpes, CNRS, Grenoble INP, LIG, France\\
\email{geompalik@hotmail.com, \{charlotte.laclau, massih-reza.amini\}@univ-grenoble-alpes.fr},
\and
Univ. Lyon, INSA-Lyon, Univ. Claude Bernard Lyon 1, UJM-Saint Etienne, CNRS, Inserm, CREATIS UMR 5220, U1206, F‐69XXX, Lyon, France\\
\email{ievgen.redko@creatis.insa-lyon.fr}}

\maketitle              

\begin{abstract}
Many information retrieval algorithms rely on the notion of a good distance that allows to efficiently compare objects of different nature. Recently, a new promising metric called Word Mover's Distance was proposed to measure the divergence between text passages. In this paper, we demonstrate that this metric can be extended to incorporate term-weighting schemes and provide more accurate and computationally efficient matching between documents using entropic regularization. We evaluate the benefits of both extensions in the task of cross-lingual document retrieval (CLDR). Our experimental results on eight CLDR problems suggest that the proposed methods achieve remarkable improvements in terms of Mean Reciprocal Rank compared to several baselines.
\end{abstract}

\section{Introduction}
Estimating distances between text passages is in the core of information retrieval applications such as document retrieval, summarization and question answering. 
Recently, in \cite{kusner2015word}, Kusner et al. proposed the Word Mover's Distance (WMD), a novel distance metric for text data. WMD is directly derived from the optimal transport (OT) theory \cite{monge81,kantorovich} and is, in fact, an implementation of the Wasserstein distance (also known as Earth Mover's distance) for textual data. 
For WMD, a source and a target text span are expressed by high-dimensional probability densities through the bag-of-words representation. 
Given the two densities, OT aims to find the map (or transport plan) that minimizes the total transportation cost given a ground metric for transferring the first density to the second. The ground metric for text data can be estimated using word embeddings \cite{kusner2015word}. 

One interesting feature of the Wasserstein distance is that it defines a proper metric on a space of probability measures. This distance presents several advantages when compared to other statistical distance measures, such as, for instance, the $f$- and the Jensen-Shannon divergences: (1) it is parametrized by the ground metric that offers the flexibility in adapting it to various data types; (2) it is known to be a very efficient metric due to its ability of taking into account the geometry of the data through the pairwise distances between the distributions' points. 
For all these reasons, the Wasserstein distance is of increasing interest to the machine learning community for various applications like: computer vision \cite{Rubner:2000:EMD:365875.365881}, domain adaptation \cite{courty:hal-01018698}, 
and clustering \cite{laclau_17}.

In this paper, our goal is to show how information retrieval (IR) applications can benefit from the Wasserstein distance. We
demonstrate that for text applications, the Wasserstein distance can naturally incorporate different weighing schemes that are particularly efficient in IR applications, such as the inverse document frequency. This presents an important advantage compared to uniform weighting considered in the previous works on the subject. Further, we propose to use the regularized version of OT \cite{conf/nips/Cuturi13}, which relies on entropic regularization allowing to obtain smoother, and therefore more stable results, and to solve the OT problem using the efficient Sinkhorn-Knopp matrix algorithm. From the application's perspective, we evaluate the use of Wasserstein distances in the task of Cross-Lingual Document Retrieval (CLDR) where given a query document (e.g., English Wikipedia entry for ``Dog'') one needs to retrieve its corresponding document in another language (e.g., French entry for ``Chien''). In this specific context we propose a novel strategy to handle out-of-vocabulary words based on morphological similarity.

The rest of this paper is organized as follows. In Section \ref{sec:preliminary}, we briefly present the OT problem and its entropic regularized version. Section \ref{sec:model} presents the proposed approach and investigates different scenarios with respect to (w.r.t.) the weighting schemes, the regularization and the word embeddings. Empirical evaluations, conducted in eight cross-lingual settings, are presented in Section \ref{sec:exp} and demonstrate that our approach is substantially more efficient than other strong baselines in terms of Mean Reciprocal Rank. The last section concludes the paper with a discussion of future research perspectives. 

\section{Preliminary knowledge}\label{sec:preliminary}
In this section we introduce the OT problem
\cite{kantorovich} as well as its entropic regularized version that will be later used to calculate the regularized Wasserstein distance.

\subsection{Optimal transport} 
OT theory, originally introduced in \cite{monge81} to study the problem of resource allocation, provides a powerful geometrical tool for comparing probability distributions. 

In a more formal way, given access to two sets of points $X_S = \{\bm{x}^{S}_i \in \mathbb{R}^d\}_{i=1}^{N_S}$ and $X_T = \{\bm{x}^{T}_i \in \mathbb{R}^d\}_{i=1}^{N_T}$, we construct two discrete empirical probability distributions as follows
\[
\hat{\mu}_S = \sum_{i=1}^{N_S}p^S_i\delta_{\bm{x}_i^S} \text{ and } \hat{\mu}_T = \sum_{i=1}^{N_T}p^T_i\delta_{\bm{x}_i^T},
\]
where $p^S_i$ and $p^T_i$ are probabilities associated to $\bm{x}^{S}_i$ and $\bm{x}^{T}_i$, respectively and $\delta_{\bm{x}}$ is a Dirac measure that can be interpreted as an indicator function taking value 1 at the position of $\bm{x}$ and $0$ elsewhere. For these two distributions, the Monge-Kantorovich problem consists in finding a probabilistic coupling $\gamma$ defined as a joint probability measure over 
$X_S \times X_T$
with marginals $\hat{\mu}_S$ and $\hat{\mu}_T$ that minimizes the cost of transport with respect to some metric $l:X_s \times X_t \rightarrow \mathbb{R}^+$: 
$$\min_{\gamma \in \Pi(\hat{\mu}_S, \hat{\mu}_T)}\langle A, \gamma\rangle_F$$
where $\langle \cdot \text{,} \cdot \rangle_F$ is the Frobenius dot product, $\Pi(\hat{\mu}_S, \hat{\mu}_T) = \lbrace \gamma \in \mathbb{R}^{N_S \times N_T}_+ \vert \gamma \bm{1} = \bm{p}^S, \gamma^T \bm{1} = \bm{p}^T\rbrace$ is a set of doubly stochastic matrices and $D$ is a dissimilarity matrix, \ie, $A_{ij} = l(\bm{x}_i^S,\bm{x}_j^T)$, defining the energy needed to move a probability mass from $\bm{x}_i^S$ to $\bm{x}_j^T$. This problem admits a unique solution $\gamma^*$ and defines a metric on the space of probability measures (called the Wasserstein distance) as follows:
$$W(\hat{\mu}_S, \hat{\mu}_T) = \min_{\gamma \in \Pi(\hat{\mu}_S, \hat{\mu}_T)}\langle A, \gamma\rangle_F.$$
The success of algorithms based on this distance is also due to \cite{conf/nips/Cuturi13} who introduced an entropic regularized version of optimal transport that can be optimized efficiently using matrix scaling algorithm. We present this regularization below. 

\subsection{Entropic regularization}
The idea of using entropic regularization has recently found its application to the optimal transportation problem \cite{conf/nips/Cuturi13} through the following objective function:
\begin{align*}
    \min_{\gamma \in \Pi(\hat{\mu}_S,\hat{\mu}_T)}\langle A, \gamma\rangle_F - \frac{1}{\lambda}E(\gamma).
\end{align*}
The second term $E(\gamma) = -\sum_{i,j}^{N_S,N_T} \gamma_{i,j}\log(\gamma_{i,j})$ in this equation allows to obtain smoother and more numerically stable solutions compared to the original case and converges to it at the exponential rate \cite{2015-Benamou-Bregman}. The intuition behind it is that entropic regularization allows to transport the mass from one distribution to another more or less uniformly depending on the regularization parameter $\lambda$. Furthermore, it allows to solve the optimal transportation problem efficiently using Sinkhorn-Knopp matrix scaling algorithm \cite{sinknopp_67}.  

\section{Word Mover's Distance for CLDR}\label{sec:model}
In this section, we explain the main underlying idea of our approach and show how the regularized optimal transport can be used in the cross-lingual information retrieval. We start with the formalization of our problem.

\subsection{Problem setup}
For our task, we assume access to two document collections $\mathcal{C}^{\ell_{1}}=\{\boldsymbol{d}_1^{\ell_1}, \ldots, \boldsymbol{d}_N^{\ell_1} \}$ and $\mathcal{C}^{\ell_2}=\{\boldsymbol{d}_1^{\ell_2}, \ldots, \boldsymbol{d}_M^{\ell_2} \}$, where $\boldsymbol{d}_n^{\ell_1}$ (resp. $\boldsymbol{d}_m^{\ell_2}$) is the $n$-th (resp. $m$-th) document written in language $\ell_1$ (resp. $\ell_2$). Let the vocabulary size of the two languages be denoted as $V^{\ell_1}$ and $V^{\ell_2}$. For the rest of the development, we assume to have access to dictionaries of embeddings $\boldsymbol{E}^{\ell_1}, \boldsymbol{E}^{\ell_2}$ where words from  $\ell_1$ and $\ell_2$  are projected into a \textit{shared} vector space of dimension $D$, hence $\boldsymbol{E}^{\ell_1}\in\mathbb{R}^{V^{\ell_1} \times D}, \boldsymbol{E}^{\ell_2}\in\mathbb{R}^{V^{\ell_2} \times D}$ and $E^{\ell_1}_k$, $E^{\ell_2}_j$ denote the embeddings of words $k,j$. As learning the bilingual embeddings is not the focus of this paper, any of the previously proposed methods can be used e.g., \cite{vulic2016bilingual,speer2017conceptnet}. A document consists of words and is represented using the Vector Space Model with frequencies. Hence, $\forall n, m: \boldsymbol{d}_n^{\ell_1}\in\mathbb{R}^{V^{\ell_1}}$, $\boldsymbol{d}_m^{\ell_2}\in\mathbb{R}^{V^{\ell_2}}$; 
the value  ${d}_{nj}^{\ell_1}$ (resp. ${d}_{mk}^{\ell_2}$) then represents the frequency of word $j$ (resp. $k$) in $\boldsymbol{d}_n^{\ell_1}\in\mathbb{R}^{V^{\ell_1}}$ (resp. $\boldsymbol{d}_m^{\ell_2}\in\mathbb{R}^{V^{\ell_2}}$).
Calculating the distance of words in the embedding's space is naturally achieved using the Euclidean distance with lower values meaning that words are similar between them. For the rest, we denote by $A(j,k)=\Vert E^{\ell_1}_j-E^{\ell_2}_k\Vert_2$ the Euclidean distance between the words $k$ and $j$ in the embedding's space. Our goal is to estimate the distance of $\boldsymbol{d}_n^{\ell_1},\boldsymbol{d}_m^{\ell_2}$, that are written in two languages, while taking advantage of the expressiveness of word embeddings and the Wasserstein distance.

\subsection{Proposed method}
In order to use the Wasserstein distance on documents, we consider that the documents $\boldsymbol{d}_n^{\ell_1}$ and $\boldsymbol{d}_m^{\ell_2}$ from different languages are both modeled as empirical probability distributions, \ie
$$\boldsymbol{d}_n^{\ell_1} = \sum_{j=1}^{V^{\ell_1}} \boldsymbol{w}_{nj}\delta_{d_{nj}^{\ell_1}} \text{ and } \boldsymbol{d}_m^{\ell_2} = \sum_{k=1}^{V^{\ell_2}} \boldsymbol{w}_{mk}\delta_{d_{mk}^{\ell_2}},$$
where $\boldsymbol{w}_{nj}$ and $\boldsymbol{w}_{mk}$ are probabilities associated with words $j$ and $k$ in $\boldsymbol{d}_n^{\ell_1}$ and $\boldsymbol{d}_m^{\ell_2}$, respectively. In order to increase the efficiency of optimal transport between these documents, it would be desirable to incorporate a proper weighting scheme that reflects the relative frequencies of different words appearing in a given text corpus. To this end, we use the following weighting schemes:
\begin{itemize}
\item \textit{term frequency} (\textit{tf}), that represents a document using the frequency of its word occurrences. This schema was initially proposed in \cite{kusner2015word} and corresponds to the case where $\boldsymbol{w}_{nj}=d_{nj}$ and $\boldsymbol{w}_{mk}=d_{mk}$. 
\item \textit{term frequency-inverse document frequency} (\textit{idf}), where the term frequencies are multiplied by the words' inverse document frequencies. In a collection of $N$ documents, the document frequency $df(j)$ is the number of documents in the collection containing the word $j$. A word's inverse document frequency penalizes words that occur in many documents. As commonly done, we use a smoothed version of \textit{idf}. Hence, we consider $\boldsymbol{w}_{nj}=d_{nj}\times \log \frac{N+1}{df(j)+1}$ and $\boldsymbol{w}_{mk}=d_{mk}\times \log \frac{M+1}{df(k)+1}$.
\end{itemize}

Furthermore, we use the Euclidean distance between the word embeddings of the two documents \cite{kusner2015word} as a ground metric in order to construct the matrix $A$. Now, we seek solving the following optimization problem:
\begin{align}
\min_{\gamma \in \Pi\left(\boldsymbol{d}_n^{\ell_1}, \boldsymbol{d}_m^{\ell_2}\right)}\langle A, \gamma\rangle_F.
\label{eq:optimization_problem1}
\end{align}
Given the solution $\gamma^*$ of this problem, we can calculate the Wasserstein distance between documents as
\begin{align*}
W(\boldsymbol{d}_n^{\ell_1}, \boldsymbol{d}_m^{\ell_2}) = \langle A, \gamma^*\rangle_F = \tr(A\gamma^*).
\end{align*}
As transforming the words of $\boldsymbol{d}_n^{\ell_1}$ to $\boldsymbol{d}_m^{\ell_2}$ comes with the cost $A(j,k)$, the optimization problem of Eq. \eqref{eq:optimization_problem1} translates to the minimization of the associated cumulative cost of transforming all the words. The value of the minimal cost is the distance between the documents.  Intuitively, the more similar the words between the documents are, the lower will be the costs associated to the solution of the optimization problem, which, in turn, means smaller document distances. For example, given ``the cat sits on the mat'' and its French translation ``le chat est assis sur le tapis'', the weights after stopwords filtering of ``cat'', ``sits'', ``mat'', and ``chat'', ``assis'', ``tapis'' will be $1/3$. Given high-quality embeddings, solving Eq. \eqref{eq:optimization_problem1} will converge to the one-to-one transformations ``cat-chat'', ``sits-assis'' and ``mat-tapis'', with very low cumulative cost as the paired words are similar. 

This one-to-one matching, however, can be less efficient when documents with larger vocabularies are used. In this case, every word can be potentially associated with, not a single, but several words representing its synonyms or terms often used in the same context. Furthermore, the problem of Eq. \eqref{eq:optimization_problem1} is a special case of the Earth Mover's distance \cite{rubner1998metric} 
and presents a standard Linear Programming problem that has a computation complexity of $\mathcal{O}(n^3\log(n))$. When $n$ is large, this can present a huge computational burden. Hence, it may be more beneficial to use the entropic regularization of optimal transport that allows more associations between words by increasing the entropy of the coupling matrix and can be solved faster, in linear time. Our second proposed model thus reads
\begin{align}
\min_{\gamma \in \Pi\left(\boldsymbol{d}_n^{\ell_1}, \boldsymbol{d}_m^{\ell_2}\right)}\langle A, \gamma\rangle_F - \frac{1}{\lambda}E(\gamma).
\label{eq:optimization_problem2}
\end{align}
As in the previous problem, once $\gamma^*$ is obtained, we estimate the entropic regularized Wasserstein distance (also known as Sinkhorn distance) as
\begin{align*}
W(\boldsymbol{d}_n^{\ell_1}, \boldsymbol{d}_m^{\ell_2}) = \langle A, \gamma^*\rangle_F = \tr(A\gamma^*) -   \frac{1}{\lambda}E(\gamma^*).
\end{align*}

Algorithm \ref{algo:cldr}  summarizes the CLDR process with Wasserstein distance. 
We also illustrate the effect of regularization, controlled by $\lambda$ in the OT problem of Eq. \eqref{eq:optimization_problem2}. Figure 1 presents the obtained coupling matrices and the underlying word matchings between  
the words of the example we considered above when varying $\lambda$. We project the words in 2-D space using t-SNE as our dimensionality reduction technique.\footnote{We use the Numberbatch embeddings presented in our experiments (Sec. \ref{sec:exp}). } Notice that high $\lambda$ values lead to the uniform association weights between all the words while the lowest value leads to a complete algorithm failure. For $\lambda=1$ the corresponding pairs are associated with the bold lines showing that this link is more likely than the other fading lines. Finally, $\lambda=0.1$ gives the optimal, one-to-one matching. This figure shows that entropic regularization encourages the ``soft" associations of words with different degrees of strength. Also, it highlights how OT accounts for the data geometry, as the strongest links occurs between the words that are closest it the space.  
\captionsetup[subfloat]{labelformat=empty}
\begin{figure}[t]
\centering
\subfloat[]{
\includegraphics[width = 0.95\textwidth]{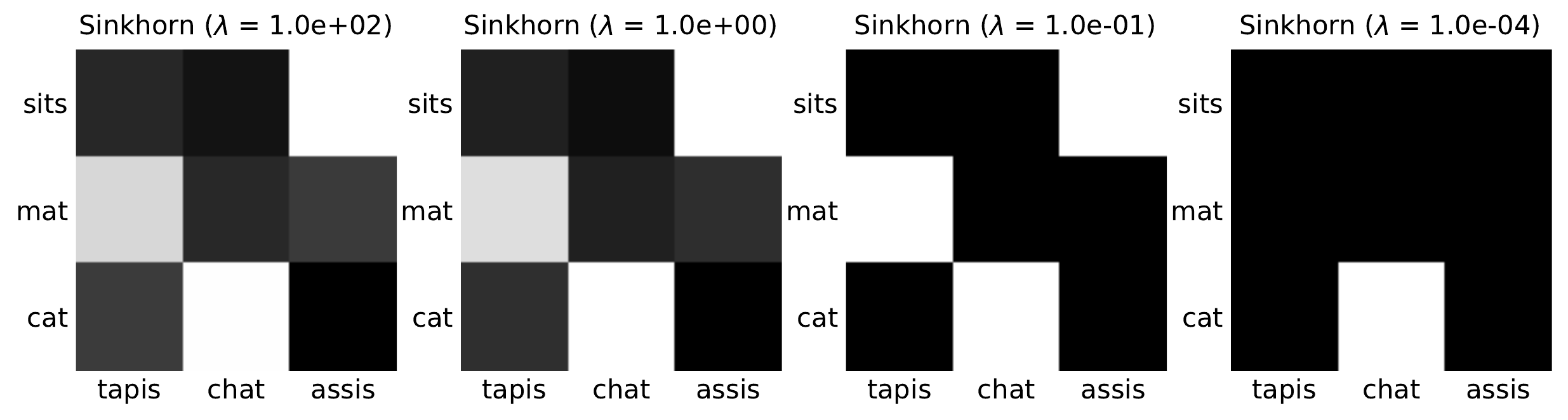}
}
\\\vspace{-0.9cm}
\subfloat[]{
\includegraphics[width = 0.95\textwidth]{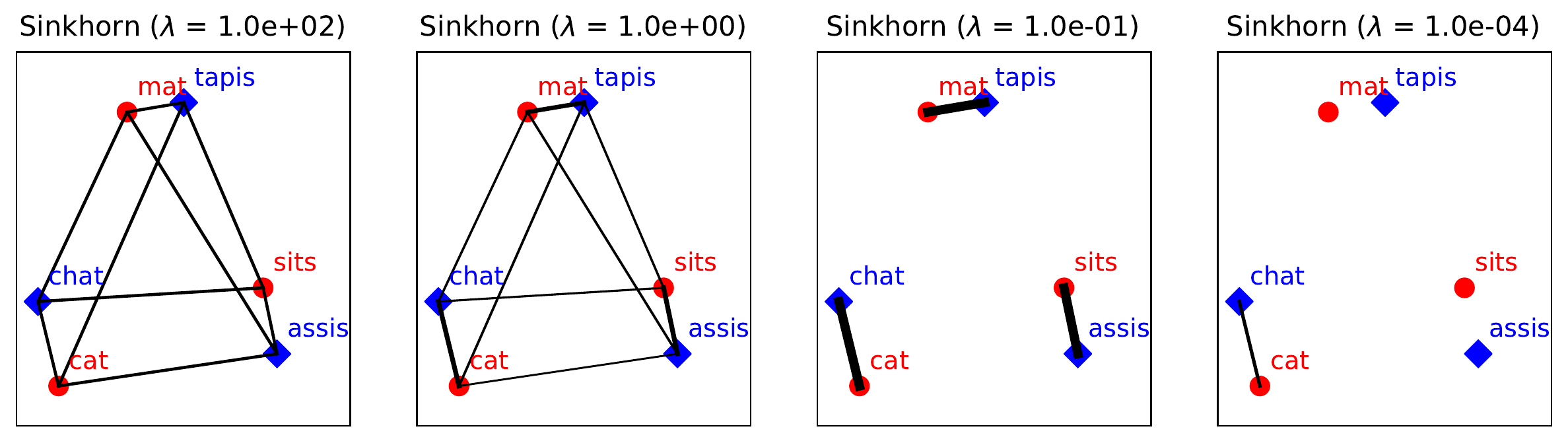}
}\vspace{-0.8cm}
\caption{The impact of entropic regularization on the transport plan. In the top matrices, whiter cells suggest stronger associations in the obtained plan. The association strength is also indicated by wider lines in the bottom figure.}
\label{fig:ot_cat}
\end{figure}

\RestyleAlgo{boxruled}
\begin{algorithm}[H]\scriptsize
\SetKwRepeat{Do}{do}{while}
 \KwData{Query $\boldsymbol{d}_n^{\ell_1}$, Corpus  $C^{\ell_2}$, Embeddings $E^{\ell_1}, E^{\ell_2}$}
\lIf{\upshape \texttt{idf}}
{
    Apply idf weights on $\boldsymbol{d}_n^{\ell_1}$, $\forall \boldsymbol{d}_m^{\ell_2} \in C^{\ell_2}$ 
}
$L_1$-normalize $\boldsymbol{d}_n^{\ell_1}$ and $\forall \boldsymbol{d}_m^{\ell_2} \in C^{\ell_2}$\;

\For{ \upshape document $\boldsymbol{d}_m^{\ell_2} \in C^{\ell_2}$}{
  \upshape 
  $dist[m] = Wass(\boldsymbol{d}_n^{\ell_1}, \boldsymbol{d}_m^{\ell_2}, A, \lambda )$ ; \texttt{\# \scriptsize *$Wass$ solves Eq. \eqref{eq:optimization_problem1} or Eq. \eqref{eq:optimization_problem2}*} 
 }
 \KwResult{$\arg\min(dist):$ increasing list of distances between $\boldsymbol{d}_n^{\ell_1}$ and $\forall \boldsymbol{d}_m^{\ell_2} \in C^{\ell_2}$ } 
\caption{CLDR with Wasserstein distance.}\label{algo:cldr}
\end{algorithm}

\subsection{Out-of-vocabulary words}\label{sec:oov}

An important limitation when using the dictionaries of embeddings $\boldsymbol{E}^{\ell_1}, \boldsymbol{E}^{\ell_2}$ is the out-of-vocabulary (OOV) words. High rates of OOV result in loss of information.
To partially overcome this limitation one needs a protocol to handle them. We propose a simple strategy for OOV that is based on the strong assumption that morphologically similar words have similar meanings. To measure similarity between words we use the Levenshtein distance, that estimates the minimum number of character edits needed to transform one word to the other. Hence, the protocol we use is as follows: in case a word $w^\ell$ is OOV, we measure its distance from every other word in $\boldsymbol{E}^{\ell}$, and select the embedding of a word whose distance is less than a threshold $t$. If there are several such words, we randomly select one. Depending on the language and on the dictionary size, this may have significant effects: for languages like English and Greek for instance one may handle OOV plural nouns, as they often formulated by adding the suffix ``s" ($\varsigma$ in Greek) in the noun (e.g., tree/trees). The same protocol can help with languages like French and German that have plenty of letters with accents such as acutes ({\'e}) and umlauts ({\"u}). 

The above strategy handles the OOV words of a language $\ell$ using its dictionary $E^\ell$. To fine-tune the available embeddings for the task of cross-lingual retrieval, we extend the argument of morphological similarity for the embeddings across languages. To achieve that, we collapse the cross-lingual embeddings of alike words. Hence, if for two words $w^{\ell_1}$ and $w^{\ell_2}$ the Levenshtein distance is zero, we use the embedding of the language with the biggest dictionary size for both. As a result, the English word ``transition" and the French word ``transition" will use the English embedding. Of course, while several words may be that similar between English and French, there will be fewer, for instance, for English and Finnish or none for English and Greek as they use different alphabets. 

We note that the assumption of morphologically similar words having similar meanings and thus embeddings is strong; we do not claim it to be anything more than a heuristic. In fact, one can come up with counter-examples where it fails. We believe, however, that for languages with less resources than English, it is an heuristic that can help overcome the high rates of OOV. Its positive or negative impact for CLDR remains to be empirically validated.


\section{Experimental framework}\label{sec:exp}

In our experiments we are interested in CLDR whose aim is to identify corresponding documents written in different languages. 
Assuming, for instance, that one has access to English and French Wikipedia documents, the goal is to identify the cross-language links between the articles. 
Traditional retrieval approaches employing bag-of-words representations perform poorly in the task as the vocabularies vary across languages, and words from different languages rarely co-occur. 
\paragraph{Datasets}
To evaluate the suitability of OT distances for cross-lingual document retrieval we 
extract four bilingual ($\ell_1-\ell_2$) Wikipedia datasets: (i) English-French, (ii) English-German, (iii) English-Greek and (iv) English-Finnish. Each dataset defines two retrieval problems: for the first ($\ell_1\rightarrow\ell_2$) the documents of $\ell_1$ are retrieved given queries in $\ell_2$; for the second ($\ell_2\rightarrow\ell_1$) the documents of $\ell_1$ are the queries. 
To construct the Wikipedia excerpts, we use the comparable Wikipedia corpora of \textit{linguatools}.\footnote{\url{http://linguatools.org/tools/corpora/wikipedia-comparable-corpora/}} 
Following \cite{vulic2015probabilistic}, the inter-language links are the golden standard and will be used to calculate the evaluation measures. 
Compared to ``ad hoc'' retrieval problems \cite{voorhees2003overview} where there are several relevant documents for each query, this is referred to as a ``known-search'' problem as there is exactly one ``right'' result for each query \cite{broder2002taxonomy}. 
Our datasets comprise 10K pairs of comparable documents; the first 500 are used for evaluating the retrieval approaches. The use of the remaining 9.5K, dubbed ``BiLDA Training'' is described in the next paragraph. 
Table \ref{tbl:data_used}(a) summarizes these datasets.
In the pre-processing steps we lowercase the documents, we remove the stopwords, the punctuation symbols and the numbers. We keep documents with more than five words and, for efficiency, we keep for each document the first 500 words. 
\begin{table}[!t]\scriptsize\centering\setlength{\tabcolsep}{3pt}
\caption{(a) Statistics for the Wikipedia datasets. W$^{\ell}$ is the corpus size in words;  V$^{\ell}$ is the vocabulary size. (b) The size of the embedding's dictionary.}\label{tbl:data_used}\vspace{-0.4cm}
\subfloat[]{
 \begin{tabular}{l cccc}\toprule
 &Wiki$_{\text{En-Fr}}$ &Wiki$_{\text{En-Ge}}$ &Wiki$_{\text{En-Fi}}$& Wiki$_{\text{En-Gr}}$\\\midrule
 $|D|$   &10K &10K &10K &10K \\
 V$^{\ell_1}$ & 33,925	&33,198	&43,230	& 34,192\\
 V$^{\ell_2}$ &26,602	&44,896	&31,112	& 28,192\\
 W$^{\ell_1}$ &36M	&36M	&58M	& 39M \\
 W$^{\ell_2}$ &24M	&31M	&15M	& 14M \\
 \bottomrule
 \end{tabular}
}
\hspace{0.4cm}
\subfloat[]{
 \begin{tabular}{l c}\toprule
Language & $|E|$ \\
\midrule
En & 417,194\\
Fr & 296,986\\
Ge & 129,405\\
Fi & 56,899 \\
Gr & 16,925 \\
 \bottomrule
 \end{tabular}
}
\end{table}
\paragraph{Systems}
One may distinguish between two families of methods for cross-lingual IR: translation-based and semantic-based. The methods of the first one use a translation mechanism to translate the query from the source language to the target language. Then, any of the known retrieval techniques can be employed. The methods of the second family project both the query and the documents of the target language in a shared semantic space. The calculation of the query-document distances is performed in this shared space. 

To demonstrate the advantages of our methods, we present results using systems that rely either on translations or on cross-lingual semantic spaces. Concerning the translation mechanism, we rely on a dictionary-based approach. To generate the
dictionaries for each language pair we use Wiktionary and, in particular, the open implementation of \cite{acs-pajkossy-kornai:2013:BUCC,CS14.864}. 
For a given word, one may have several translations: we pick the candidate translation according to a unigram language model learned on ``BiLDA training'' data.
In the rest, we compare:\footnote{We release the code at: \url{https://github.com/balikasg/WassersteinRetrieval}.}

\noindent\texttt{-tf}: Euclidean distance between the term-frequency representation of documents. To be applied for CLDR, the query needs to be translated in the language of the target documents. 

\smallskip

\noindent\texttt{-idf}: Euclidean distance between the idf representation of documents. As with \texttt{tf}, the query needs to be translated. 

\smallskip

\noindent\texttt{-nBOW$_{x}$}: \texttt{nBOW} (neural bag-of-words) represents documents by a weighted average of their words' embeddings \cite{mitchell2010composition,BlacoeL12}. If $x$=\texttt{tf} the output is the result of averaging  the embeddings of the occurring words.  If $x$=\texttt{idf}, then the embedding of each word is multiplied with the word's inverse document frequency. Having the \texttt{nBOW} representations the distances are calculated using the Euclidean distance. 
\texttt{nBOW$_{x}$} methods can be used both with cross-lingual embeddings, or with mono-lingual embeddings if the query is translated.   

\smallskip

\noindent\texttt{-BiLDA}: Previous work found the bilingual Latent Dirichlet Allocation (BiLDA) 
to yield state-of-the-art results, we cite for instance \cite{fukumasu2012symmetric,vulic2015probabilistic,wang2016cross}. 
BiLDA is trained on comparable corpora and learns aligned per-word topic distributions between two or more languages. During inference it projects unseen documents in the \textit{shared} topic space where cross-lingual distances can be calculated efficiently. We train BiLDA separately for each language pair with 300 topics.  We use collapsed Gibbs sampling for inference \cite{vulic2015probabilistic}, which we implemented with Numpy \cite{walt2011numpy}. Following previous work, we set the Dirichlet hyper-parameters $\alpha=50/K$ ($K$ being the number of topics) and $\beta=.01$. We let 200 Gibbs sampling iterations for burn-in and then sample the document distributions every 25 iterations until the $500$-th Gibbs iteration. For learning the topics, we used the ``BiLDA training'' data. Having the per-document representations in the shared space of the topics, we use entropy as distance, following \cite{fukumasu2012symmetric}.

\smallskip

\noindent\texttt{-Wass$_{x}$}: is the proposed metric given by Eq. \ref{eq:optimization_problem1}. If $x$=\texttt{tf}, it is equivalent to that of \cite{kusner2015word}, as for generating the high-dimensional source and target histograms the terms' frequencies are used.  
If $x$=\texttt{idf}, the idf weighting scheme is applied. 

\smallskip

\noindent\texttt{-Entro\_Wass$_{x}$} is the proposed metric given by Eq. \ref{eq:optimization_problem2}. The subscript $\boldsymbol{x}$ reads the same as for the previous approach.
We implemented \texttt{Wass$_{x}$} and \texttt{Entro\_Wass$_{x}$} with Scikit-learn \cite{sklearn} using the solvers of POT \cite{flamary2017pot}.\footnote{For \texttt{Entro\_Wass} we used the \texttt{sinkhorn2} function with \texttt{reg=0.1, numItermax=50, method='sinkhorn\_stabilized'} arguments to prevent numerical errors.} For the importance of the regularization term in Eq. \eqref{eq:optimization_problem2}, we performed grid-search  $\lambda\in\{10^{-3}, \ldots, 10^2\}$ and found $\lambda=0.1$  to consistently perform the best.

For the systems that require embeddings (\texttt{nBOW}, \texttt{Wass}, \texttt{Entro\_Wass}), we use the Numberbatch pre-trained embeddings of \cite{speer2017conceptnet}.\footnote{The v17.06 vectors: \url{https://github.com/commonsense/conceptnet-numberbatch}}
The Numberbatch embeddings are 300-dimensional embeddings for 78 languages, that project words and short expressions of these languages in the same shared space and were shown to achieve state-of-the-art results in cross-lingual tasks \cite{speer2017conceptnet,speer2017conceptnetSemeval}.  

Complementary to \texttt{tf} and \texttt{idf} document representations, we also evaluate the heuristic we proposed for OOV words in Section \ref{sec:oov}. We select the threshold $t$ for the Levenhstein distances to be 1, and we denote with \texttt{tf+} and \texttt{idf+} the settings where the proposed OOV strategy is employed. 
\begin{table*}[!t]\setlength{\tabcolsep}{3pt}\scriptsize  \centering
\caption{The MRR scores achieved by the systems. With asterisks we to denote our original contributions. \texttt{Entro\_Wass$_{\text{idf+}}$} consistently achieves the best performance by a large margin. }\label{tbl:results}
 \begin{tabular}{l cc cc cc cc }
 \toprule
& En$\rightarrow$ Fr  & Fr$\rightarrow$ En &  En$\rightarrow$ Ge & Ge$\rightarrow$ En &En$\rightarrow$ Gr & Gr$\rightarrow$ En & En $\rightarrow$ Fi & Fi$\rightarrow$ En\\
  \midrule
  \multicolumn{9}{c}{Systems that rely on topic models} \\\cmidrule(lr){2-9}
\texttt{BiLDA} &  .559	&	.468	&	.560	&	.536	&	.559	&	.508	&	.622	&	.436	\\ 
  \midrule
  \multicolumn{9}{c}{Systems that rely on translations} \\
\cmidrule(lr){2-9}

\texttt{tf} 			   &  .421 & .278 & .287 & .433 & .441 & .054 & .068 & .025 \\ 
\texttt{idf} 			   &.575 & .433 & .478 & .516 & .535 & .126 & .081 & .028 \\
\texttt{nBOW$_{\text{tf}}$}	   &.404 & .276 & .378 & .440 & .531 & .132 & .329 & .042 \\
\texttt{nBOW$_{\text{idf}}$}	   &.550 & .488 & .449 & .509 & .577 & .256 & .398 & .081 \\

\texttt{Wass$_{\text{tf}}$}        &.704 & .691 & .620 & .655 & .656 & .269 & .424 & .092 \\
\texttt{Wass$_{\text{idf}}$}*       &.748 & .766 & .678 & .706 & .687 & .463 & .412 & .163 \\

\texttt{Entro\_Wass$_{\text{tf}}$}* &.692 & .706 & .615 & .666 & .640 & .262 & .422 & .089 \\
\texttt{Entro\_Wass$_{\text{idf}}$}*&.745 & .794 & .675 & .720 & .683 & .467 & .425 & .171 \\

\midrule
  \multicolumn{9}{c}{Systems that rely on cross-lingual embeddings} \\
\cmidrule(lr){2-9}

\texttt{nBOW$_{\text{tf}}$} 	&.530 & .490 & .493 & .464 & .237 & .121 & .449 & .217\\
\texttt{nBOW$_{\text{idf}}$}	&.574 & .546 & .521 & .502 & .341 & .179 & .470 & .267 \\

\texttt{Wass$_{\text{tf}}$}    &.744 & .748 & .660 & .681 & .404 & .407 & .582 & .434  \\
\texttt{Wass$_{\text{idf}}$}*   &.778 & .784 & .703 & .718 & .507 & .465 & .620 & .479 \\ 
\texttt{Entro\_Wass$_{\text{tf}}$}* 	& .753 & .786 & .677 & .710 & .424 & .494 & .582 & .607  \\
\texttt{Entro\_Wass$_{\text{idf}}$}*  	&.799 & .820 & .717 & .756 & .523 & .549 & .620 & .643 \\\midrule 

  \multicolumn{9}{c}{Handling OOV words \& cross-lingual embeddings} \\\cmidrule(lr){2-9}
\texttt{nBOW$_{\text{tf+}}$}&  .518 & .541 & .426 & .468 & .193 & .148 & .635 & .451  \\ 
\texttt{nBOW$_{\text{idf+}}$}& .653 & .659 & .597 & .592 & .407 & .349 & .693 & .544 \\ 
\texttt{Wass$_{\text{tf+}}$}&  .815 & .836 & .788 & .801 & .675 & .435 & .845 & .731  \\ 
\texttt{Wass$_{\text{idf+}}$}* &.867 & .869 & .812 & .837 & .721 & .599 & .856 & .786 \\ 
\texttt{Entro\_Wass$_{\text{tf+}}$}* 	&.830 & .856 & .796 & .812 & .718 & .555 & .851 & .816 \\ 
\texttt{Entro\_Wass$_{\text{idf+}}$}* 	&\textbf{.875} & \textbf{.887} & \textbf{.828} & \textbf{.855 }& \textbf{.741} & \textbf{.695} & \textbf{.864} & \textbf{.850} \\ 
\bottomrule
 \end{tabular}
\end{table*}

\paragraph{Results} 
As evaluation measure, we report the Mean Reciprocal Rank (MRR) \cite{voorhees1999trec} which accounts for the rank of the correct answer in the returned documents. Higher values signify that the golden documents are ranked higher. Table \ref{tbl:results} presents the achieved scores for the CLDR problems. 

There are several observations from the results of this table. First, notice that the results clearly establish the superiority of the Wasserstein distance for CLDR. Independently of the representation used (\texttt{tf}, \texttt{idf}, \texttt{tf+}, \texttt{idf+}) the performance when the Wasserstein distance is used is substantially better than the other baselines. This is due to the fact that the proposed distances account for the geometry of the data. In this sense, they essentially implement optimal word-alignment algorithms as the calculated transportation cost uses the word representations in order to minimize their transformation from the source to the target document. Although \texttt{nBOW} also uses exactly the same embeddings, it performs a weighted averaging operation that results in information loss.

Comparing the two proposed methods, we notice that the approach with the entropic regularization (\texttt{Entro\_Wass}) outperforms in most of the cases its original version \texttt{Wass}. This suggests that using regularization in the OT problem improves the performance for our application. As a result, using \texttt{Entro\_Wass} is not only faster and GPU-friendly, but also more accurate. Also, both approaches consistently benefit from the \texttt{idf} weighting scheme. The rest of the baselines, although competitive, perform worse than the proposed approaches. 

Another interesting insight stems from the comparison of the translation-based and the semantic-based approaches. The results suggest that the semantic-based approaches that use the Numberbatch embeddings perform better, meaning that the machine translation method we employed introduces more error than the imperfect induction of the embedding spaces. This is also evident by the performance decrease of \texttt{tf} and \texttt{idf} when moving from language pairs with more resources like ``En-Fr'' to more resource deprived pairs like ``En-Fi'' or ``En-Gr''.
While one may argue that better results can be achieved with a better-performing translation mechanism, the important outcome of our comparison is that both families of approaches improve when the Wasserstein distance is used. Notice, for instance, the characteristic example of the translation based systems for ``Fi$\rightarrow$En'': \texttt{tf}, \texttt{idf} and their \texttt{nBOW} variants perform poorly (MRR $\le$ 0.09), suggesting low-quality translations; still \texttt{Wass} and \texttt{Entro\_Wass} achieve remarkable improvements (MRR$\sim0.17$), using the same resources. 

Our last comments concern the effect of the OOV protocol. Overall, having such a protocol in place benefits \texttt{Wass} and \texttt{Entro\_Wass} as the comparison of the \texttt{tf} and \texttt{idf} with \texttt{tf+} and \texttt{idf+} variants suggests. The impact of the heuristic is more evident for the ``En-Gr'' and ``En-Fi'' problems that gain several ($\sim$.20) points in terms of MRR. This is also due to the fact that the proposed OOV mechanism reduces the OOV rates as Greek and Finnish have the smallest embeddings dictionary as shown in Table \ref{tbl:data_used}b.

\section{Conclusions}\label{sec:conclusion}

In this paper, we demonstrated that the Wasserstein distance and its regularized version  naturally incorporate term-weighting schemes. We also proposed a novel protocol to handle OOV words based on morphological similarity. Our experiments, carried on eight CLDR datasets, established the superiority of the Wasserstein distance compared to other approaches as well as the interest of integrating entropic regularization to the optimization, and $tf-idf$ coefficients to the word embeddings. Finally, we showed the benefits of our OOV strategy, especially when the size of the embedding's dictionary for a language is small.

Our study opens several avenues for future research. First, we plan to evaluate the generalization of the Wasserstein distances for ad hoc retrieval, using for instance  the benchmarks of the CLEF ad hoc news test suites. 
Further, while we showed that entropic regularization greatly improves the achieved results, it remains to be studied how one can apply other types of regularization to the OT problem. For instance, one could expect that group sparsity inducing regularization applied in the CLDR context can be a promising direction as semantically close words intrinsically form clusters and thus it appears meaningful to encourage the transport within them. Lastly, CLDR with Wasserstein distances is an interesting setting for comparing methods for deriving cross-lingual embeddings as their quality directly impacts the performance on the task. 


\vspace{-0.3cm}

\bibliographystyle{unsrt}
\bibliographystyle{splncs03}
\bibliography{emnlp2017}
\end{document}